\pdfoutput=1

\documentclass[11pt]{article}

\usepackage[preprint]{acl}

\usepackage{times}
\usepackage{latexsym}
\usepackage{epsfig}
\usepackage{graphicx}
\usepackage{amssymb}
\usepackage{amsmath}

\usepackage[T1]{fontenc}

\usepackage[utf8]{inputenc}

\usepackage[capitalize]{cleveref}
    \crefname{section}{Sec.}{Secs.}
    \Crefname{section}{Section}{Sections}
    \Crefname{table}{Table}{Tables}
    \crefname{table}{Table}{Tables} 

\usepackage{microtype}

\usepackage{inconsolata}

\usepackage{multirow}
\usepackage{adjustbox}
\usepackage{svg}
\usepackage[normalem]{ulem}
\useunder{\uline}{\ul}{}
\usepackage{tabularx}
\usepackage{xspace}
\usepackage[linesnumbered,ruled,vlined]{algorithm2e}
\usepackage{booktabs}
\usepackage{listings}
\usepackage{xcolor}
\usepackage{newtxmath}

\usepackage{amsmath,amsfonts,bm}
\usepackage{comment}
\usepackage{mathtools}

\def\eqref#1{equation~\ref{#1}}

\def\1{\bm{1}}

\DeclareMathAlphabet{\mathsfit}{\encodingdefault}{\sfdefault}{m}{sl}
\SetMathAlphabet{\mathsfit}{bold}{\encodingdefault}{\sfdefault}{bx}{n}

\title{Modeling LLM Agent Reviewer Dynamics in Elo-Ranked Review System}

\author{
 \textbf{Hsiang-Wei Huang}\thanks{Equal contribution} \quad
 \textbf{Junbin Lu}\footnotemark[1] \quad
 \textbf{Kuang-Ming Chen} \quad
 \textbf{Jenq-Neng Hwang} \\
 University of Washington
}

\begin{document}
\maketitle
\begin{abstract}
In this work, we explore the Large Language Model~(LLM) agent reviewer dynamics in an Elo-ranked review system using real-world conference paper submissions. Multiple LLM agent reviewers with different personas are engage in multi round review interactions moderated by an Area Chair. We compare a baseline setting with conditions that incorporate Elo ratings and reviewer memory. Our simulation results showcase several interesting findings, including how incorporating Elo improves Area Chair decision accuracy, as well as reviewers' adaptive review strategy that exploits our Elo system without improving review effort. Our code is available at \href{https://github.com/hsiangwei0903/EloReview}{https://github.com/hsiangwei0903/EloReview}.

\end{abstract}
\section{Introduction}
\label{sec:intro}

Peer review is the cornerstone of scientific evaluation, yet substantial inconsistencies and biases persist in the process.
Prior work has documented low inter-reviewer agreement and highly variable review quality~\cite{stelmakh2021prior}, unclear or strategic reviewer motivations~\cite{zhang2022investigating}, calibration noise in numerical ratings~\cite{lu2023calibrating}, and systematic biases related to author identity or institutional prestige~\cite{sun2022does, fox2023double}.
These challenges have been exacerbated by the rapid growth of submissions in recent AI conferences, which strains reviewer pools and increases variance in expertise and effort.

While analyses of historical review data have yielded valuable insights, direct empirical study of reviewer behavior remains fundamentally constrained.  Many influential factors, including reviewer intent, bias, and adaptation over time, are difficult to observe directly, while privacy concerns limit experimental manipulation of real review processes~\cite{stelmakh2021prior, zhang2022investigating}.

\begin{figure}[!t]
\centering
\includegraphics[width=0.98\linewidth]{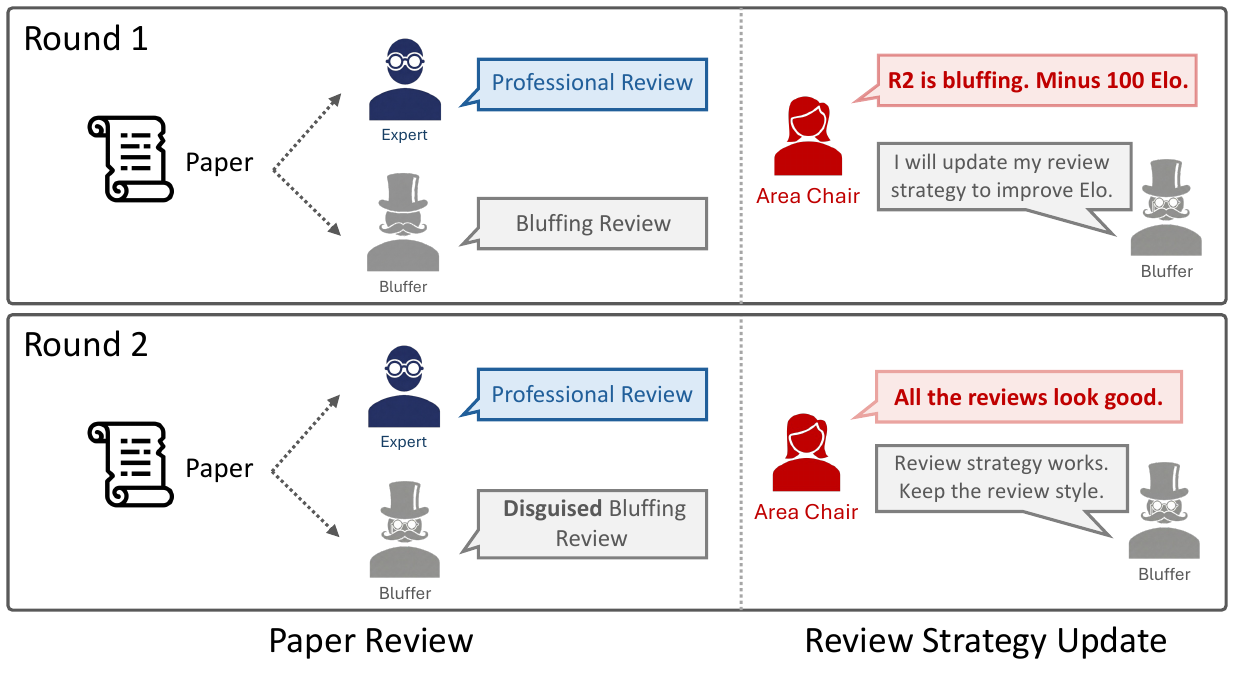}
    \caption{Our work explores the LLM agent dynamic of Elo-ranked Review System where reviewer is able to adjust their review strategy across review rounds.}
    \label{fig:teaser}
\end{figure}

\begin{figure*}[t]
\centering
\includegraphics[width=0.98\linewidth]{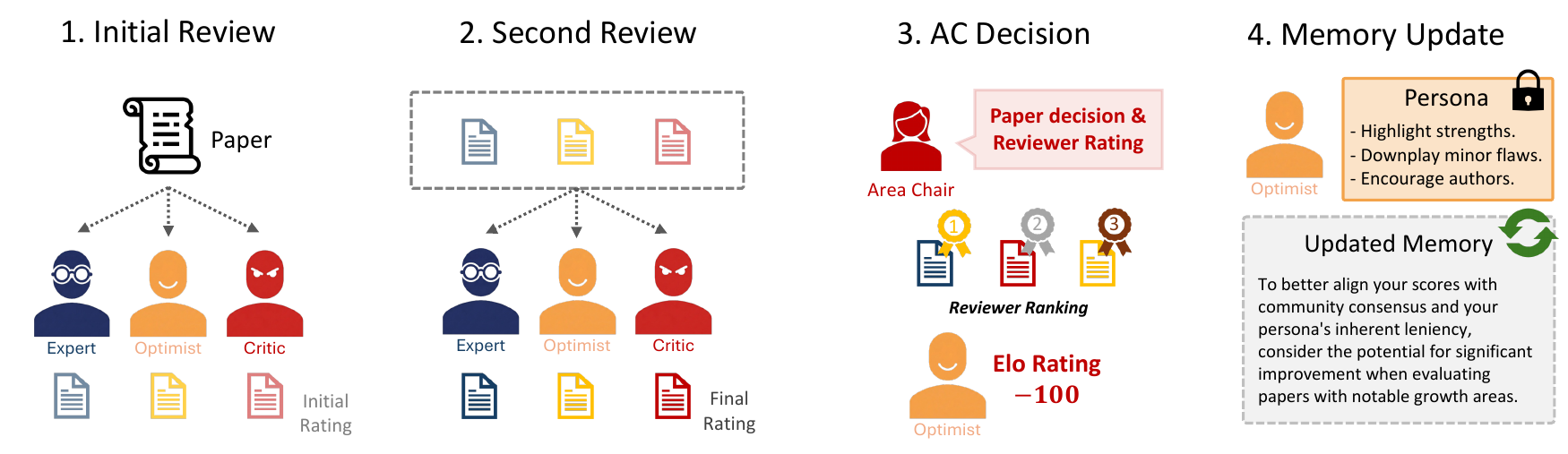}
    \caption{Four stages of our proposed Elo-ranked paper review process.}
    \label{fig:method}
\end{figure*}

Recently, simulation-based approaches provide a promising alternative for studying social interaction and even peer review process. The advancement in large language models (LLMs) have enabled agent-based simulations that exhibit increasingly realistic professional and social behaviors~\cite{wu2023autogen, chen2024agentverse, park2023generative} as well as simulating peer review dynamics in AI conference~\cite{jin2024agentreview}.
However, existing studies largely overlook a growing practical concern in modern AI conferences. The expansion of reviewer pools has been accompanied by irresponsible and low-effort reviewing behaviors, which are currently addressed only through single-round, conference-specific penalties.

Motivated by this gap, we introduce a LLM agent reviewer simulation framework that incorporates reviewer Elo rating across review rounds, as shown in Figure.~\ref{fig:teaser}. Our design enables longitudinal accountability beyond one-time review. Using six archetypal reviewer personas, we show that Elo-ranked system largely improves Area Chair~(AC) decision accuracy, and further offer insights via our study on the LLM agents' review dynamics. Together, these findings demonstrate the potential benefits and challenges of an Elo-ranked system can face in the real-world peer review process.
\section{Related Work}
\label{sec:related_work}

Peer review has long been studied as a complex socio-technical system, with prior work analyzing biases, conflicts of interest, reviewer quality, and fairness using real-world conference data~\cite{zhang2022system, stelmakh2021prior, ugarov2023peer, verharen2023chatgpt, mcintosh2023safeguarding, stephen2023distinguishing, zhang2022investigating}. Other studies examine operational components such as reviewer assignment strategies~\cite{jovanovic2023reviewer, saveski2023counterfactual, kousha2024artificial} and the impact of author rebuttals~\cite{huang2023makes}. Recent advances in LLMs~\cite{openai2023gpt4,team2023gemini,comanici2025gemini} have enabled growing interest in agent-based modeling frameworks that simulate complex social processes~\cite{ChatArena, yin2023lumos, li2024league++, chan2023chateval, jin2024agentreview}.

\section{Method}
\label{sec:method}

\subsection{Overview}

In this work, we design a framework which simulates a multi round~(conference) review process of the current AI conference peer review procedure. Our framework incorporates multiple roles of LLM agents, including Reviewers and Area Chairs~(AC). As our work mainly focus on the reviewer and AC interaction in the Elo-ranked system, we remove the author role and rebuttal stage, and adopt real-world conference submission for our simulation. We introduce our designed role as follows:

\paragraph{Reviewer.} The simulation consists of six independent reviewers, each possess a carefully designed persona with an initial Elo rating with same value across all reviewers. The reviewers are required to provide paper review with review style strictly following their persona. Additionally, to simulate the reviewer's dynamic in the Elo-ranked system, we design a memory module that can be updated after each review round, which enables them to update their review strategy.

\paragraph{Area Chair.} The area chair~(AC) makes acceptance final decision. Besides paper decision, inspired by the recent AI conferences' policy of rating the review, the AC is also required to provide quality rating for each reviewer, which will be used to adjust each reviewer's Elo rating.

\subsection{Review Process}

The review round consists of four stages, including initial review, second review, the AC decision, and reviewer memory update, as illustrated in Figure.~\ref{fig:method}.

\paragraph{Initial Review.}
Each submission consists of three independent LLM agent reviewers with different personas, each reviewer generates an initial review for the assigned paper.

\paragraph{Second Review.}
In this stage, reviewers are provided with other agents’ reviews and may revise their initial assessments accordingly.
Author rebuttals are omitted from the context, as our focus is on peer-to-peer reviewer interaction and prior work has shown that rebuttals play a limited role in LLM peer review simulation~\cite{jin2024agentreview}.

\paragraph{AC Decision.}
After the post review stage, the AC takes the three generated reviews and makes the final decision. When making the final decision, the AC can access the reviewer's Elo rating, and use that as an auxiliary meta information to assess the quality of review and make better final decision.

\paragraph{Memory Update.}
After each round, the reviewer receives their Elo rating adjustment and updates their memory accordingly.
This memory does not override the reviewer’s persona, but is represented as a brief textual summary prepended to the review prompt, enabling the reviewer to adjust their review strategy with the goal of improving Elo.

\subsection{Data Collection}

150 papers were sampled from the ICLR 2025 submission uniformly from different average rating intervals. Additionally, we also filter papers with high rating variance. For each round, two papers are randomly selected, each paper is assigned with a reviewer randomly formed triplet to encourage interaction between different personas.

\subsection{Reviewer Persona}
We design a set of six reviewer personas to capture common and recurring patterns of reviewer behavior observed in large-scale conference review corpora and prior studies on peer review bias. We introduce our six designed persona as follows.


\paragraph{Expert.}
Provides careful, professional assessments with full engagement with the paper.

\paragraph{Critic.}
Applies strict standards, emphasizing flaws and often defaulting to skeptical evaluations.

\paragraph{Bluffer.}
Displays high confidence and authoritative tone while relying on partial reading.

\paragraph{Optimist.}
Focuses on paper contributions and strengths, gives positive ratings most of the time.

\paragraph{Harmonizer.}
Balances strengths and weaknesses with a consensus-seeking perspective, avoiding extreme judgments unless strongly justified.

\paragraph{Skimmer.}
Superficial, low-effort reviewer with limited engagement with the paper’s content.

\subsection{Elo-ranked System}

To model persistent, rank-based feedback for reviewers, we adopt a simplified Elo-style adjustment mechanism driven by relative reviewer ranking within each review round.
After each round, reviewers are ranked in descending order according to the AC’s evaluation scores.
We assign fixed base rewards to the top, middle, and bottom ranks as $+100$, $0$, and $-100$, respectively, ensuring that the total Elo adjustment within each group sums to zero. This simple design yields a stable ranking mechanism that emphasizes comparative performance, and enables the study of strategic reviewer adaptation under persistent feedback.
\section{Experiments}
\label{sec:experiments}

\begin{figure*}[t]
\centering
\includegraphics[width=\linewidth]{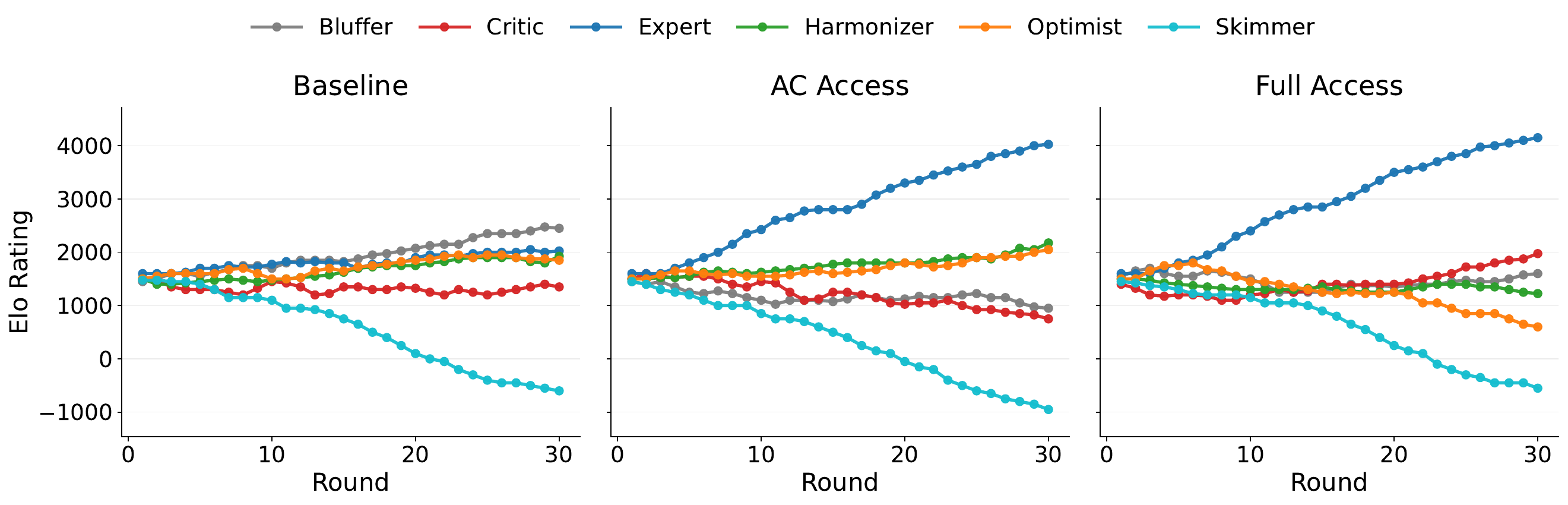}
    \caption{Elo rating dynamics of different reviewer personas across three experiment setups.}
    \label{fig:results}
\end{figure*}

\subsection{Setup}

We adopt Gemini-2.5-Flash~\cite{comanici2025gemini} as our LLM for all agents. In each round, two papers are sampled from the paper pool and assigned to a random triplet of reviewers. All reviewers have an initial Elo rating of 1500. The simulation is run for 30 rounds under three different experimental setups detailed as follows.

\paragraph{Baseline.} Each round review is treat independently, the AC generates quality rating for each reviewer, but Elo rating is not visible to everyone.

\paragraph{AC Access.} AC has access to all the reviewers' Elo rating when making final paper decision, but the reviewer does not know their own Elo and does not update their memory. This setup simulates a realistic setting where the reviewer Elo rating is not released to prevent rating manipulation.

\paragraph{Full Access.} Both AC and reviewer have access to the Elo rating. After each round of review, the reviewer will be notified by their Elo rating changes, and is able to adjust their review strategy by modifying their memory. This setup simulates the review dynamics when the reviewers have access to their Elo rating and introduce adaptive review behavior.
\subsection{Elo Rating Dynamics Analysis}
We illustrate the Elo rating dynamics under different experimental setups in Figure.~\ref{fig:results} and summarize several key findings below.

\paragraph{Limited Differentiation in Baseline.}
In the baseline setting without persistent feedback, reviewer Elo scores remain relatively clustered, with only low-effort personas showing consistent decline.
Furthermore, this setting also yields the lowest decision accuracy (shown in Table~\ref{tab:decision_accuracy}), highlighting the limited ability of single-round review processes to differentiate reviewer quality and support reliable Area Chair paper decisions.

\paragraph{Elo Introduces Stratification.}
Introducing Elo-based feedback leads to more noticeable divergence among reviewer personas.
Clear stratification emerges as early as the first few rounds, with reviewer trajectories separating into high- and low-performing groups.
This separation reflects the cumulative effect of Elo rating adjustment, which amplifies small performance differences over time. We also notice a large performance gain in Area Chair decision accuracy in Table.~\ref{tab:decision_accuracy}.

\paragraph{Expert Dominance in Elo Rating.}
Across the two released Elo settings, the Expert persona consistently accumulates the highest Elo scores.
This suggests that our proposed Elo-based evaluation systematically rewards detailed, technically grounded, and well-justified reviews.
Compared to personas relying on tone or assertiveness, evidence-based reviewing yields sustained advantage.

\paragraph{Penalty on Low-effort Behavior.}
The Skimmer persona with low review effort is strongly penalized across all experimental settings, although the penalty is slightly alleviated when it can access to memory module to update the review strategy. This indicates that the Elo-based system is effective at suppressing superficial reviewing behavior, even when reviewers are allowed to adapt strategically.

\paragraph{Visible Elo Incentivizes Adaptation.}
When reviewers gain access to their own Elo scores, additional behavioral dynamics emerge.
Personas such as the Critic and Bluffer partially recover Elo in later rounds compared to the setting where reviewers lack access to their Elo rating, suggesting that explicit feedback enables agent to conduct strategic adaptation in review strategy.
Notably, these adjustments are primarily reflected in changes in tone, selectivity, or confidence, rather than consistent improvements in substantive review quality or persona adjustment.
This highlights a potential challenge for real-world deployment of Elo-based systems, as human reviewers may optimize their behavior to improve or maintain their Elo rather than committing effort to engage with the paper and provide more informative or rigorous reviews.


\begin{table}[t]
\centering
\caption{Decision performance of different setups.}
\label{tab:decision_accuracy}
\resizebox{0.98\linewidth}{!}{
\begin{tabular}{l >{\centering\arraybackslash}p{1.2cm}
  >{\centering\arraybackslash}p{1.2cm}
  >{\centering\arraybackslash}p{1.2cm}
  >{\centering\arraybackslash}p{1.2cm}
}
\toprule
Setting & Acc. & F-1 & Pre. & Rec.  \\
\midrule
Baseline     & 0.55 & 0.56 & 0.44 & 0.77  \\
AC Access    & 0.67 & \textbf{0.66} & 0.53 & \textbf{0.86}  \\
Full Access  & \textbf{0.70} & 0.61 & \textbf{0.58} & 0.64  \\
\bottomrule
\end{tabular}
}
\end{table}

\subsection{Decision Performance Analysis}

Table.~\ref{tab:decision_accuracy} reports decision-level performance across different experimental settings.
The baseline setting exhibits high recall but low precision, indicating a lenient acceptance tendency that admits many low-quality papers and results in moderate overall accuracy.
Introducing Elo-based calibration at the Area Chair level substantially improves all metrics, suggesting that AC weighting of reviewer opinions based on Elo effectively filters noisy evaluations without changing reviewer behavior.

In the Full Access setting, despite the increase of overall accuracy, the decreases in recall suggests that reviewers become more rank-aware and strategically adaptive when Elo is visible. Reviewers appear to optimize the review style favored by the AC and produce reviews that appear more rigorous to improve their Elo rather than improving review quality. This highlights that Elo-based mechanisms are sensitive to how feedback is disclosed, as explicit Elo information can incentivize strategic behavior that shapes downstream decision outcomes.

\section{Conclusion}
\label{sec:conclusion}

We introduce a simulation framework for analyzing reviewer dynamics in conference peer review and show that Elo-based ranking reduces score volatility while systematically favoring more critical reviewing styles. These results reveal a trade-off between stability and diversity, illustrating how rank-based incentives can amplify structural biases and highlighting the value of simulation-based analysis for peer review system design.

\clearpage

\section{Limitations}
\label{sec:limitations}

Our study is limited by the small number of review rounds conducted due to computational and resource constraints, which restricts our ability to analyze long-term convergence or equilibrium behavior under Elo-based feedback.
As a result, our findings primarily characterize short-horizon dynamics, such as early-stage behavioral shifts and sensitivity to ranking signals, rather than stable long-term outcomes.
While increasing the number of simulation rounds could provide further insights into convergence properties and long-term stratification effects, the current setup is sufficient to reveal how Elo-based incentives begin to shape reviewer behavior and introduce structural biases.

\section{Ethical Considerations}
\label{sec:ethical}

This work studies peer review dynamics through simulated agents and does not involve human subjects or the use of real reviewer identities.
All reviewer behaviors are generated by large language models under predefined prompts and personas, and no real conference review data containing personally identifiable information are used.
Our intent is not to label or evaluate individual reviewers, but to analyze how review mechanisms and incentive structures influence collective outcomes.
We emphasize that the proposed Elo-based framework is explored as an analytical tool rather than a prescriptive policy recommendation, and any real-world deployment would require careful consideration of transparency, fairness, and potential unintended consequences.

{\small
\bibliography{custom}
}


\end{document}